\documentclass[runningheads]{llncs}
\usepackage{graphicx}
\usepackage{comment}
\usepackage{amsmath,amssymb} 
\usepackage{color}
\usepackage{times}
\usepackage{epsfig}
\usepackage{multirow}
\usepackage{float}
\usepackage{booktabs}
\usepackage{url}

\usepackage[width=122mm,left=12mm,paperwidth=146mm,height=193mm,top=12mm,paperheight=217mm]{geometry}

\begin{document}
\mainmatter
\def\ECCVSubNumber{1928}  

\title{Attribute-aware Identity-hard Triplet Loss for Video-based Person Re-identification} 

\titlerunning{Attribute-aware Identity-hard Triplet Loss for Video-based Person Re-identification}
%
\author{Zhiyuan Chen\inst{1} \and
Annan Li\inst{1} \and
Shilu Jiang\inst{1} \and
Yunhong Wang\inst{1}}
\authorrunning{Z. Chen et al.}
%
\institute{School of Computer Science and Engineering, Beihang University, Beijing, China
	\email{\{dechen,liannan,shilu\_jiang,yhwang\}@buaa.edu.cn}}
\maketitle

\begin{abstract}
	Video-based person re-identification (Re-ID) is an important computer vision task. The batch-hard triplet loss frequently used in video-based person Re-ID suffers from the Distance Variance among Different Positives (DVDP) problem. In this paper, we address this issue by introducing a new metric learning method called Attribute-aware Identity-hard Triplet Loss (AITL), which reduces the intra-class variation among positive samples via calculating attribute distance. To achieve a complete model of video-based person Re-ID, a multi-task framework with Attribute-driven Spatio-Temporal Attention (ASTA) mechanism is also proposed. Extensive experiments on MARS and DukeMTMC-VID datasets shows that both the AITL and ASTA are very effective. Enhanced by them, even a simple light-weighted video-based person Re-ID baseline can outperform existing state-of-the-art approaches. The codes has been published on \url{https://github.com/yuange250/Video-based-person-ReID-with-Attribute-information}.
	
\end{abstract}

\section{Introduction}
\label{sec:introduction}

\begin{figure}[t]
	\centering
	\includegraphics[width=0.95\textwidth]{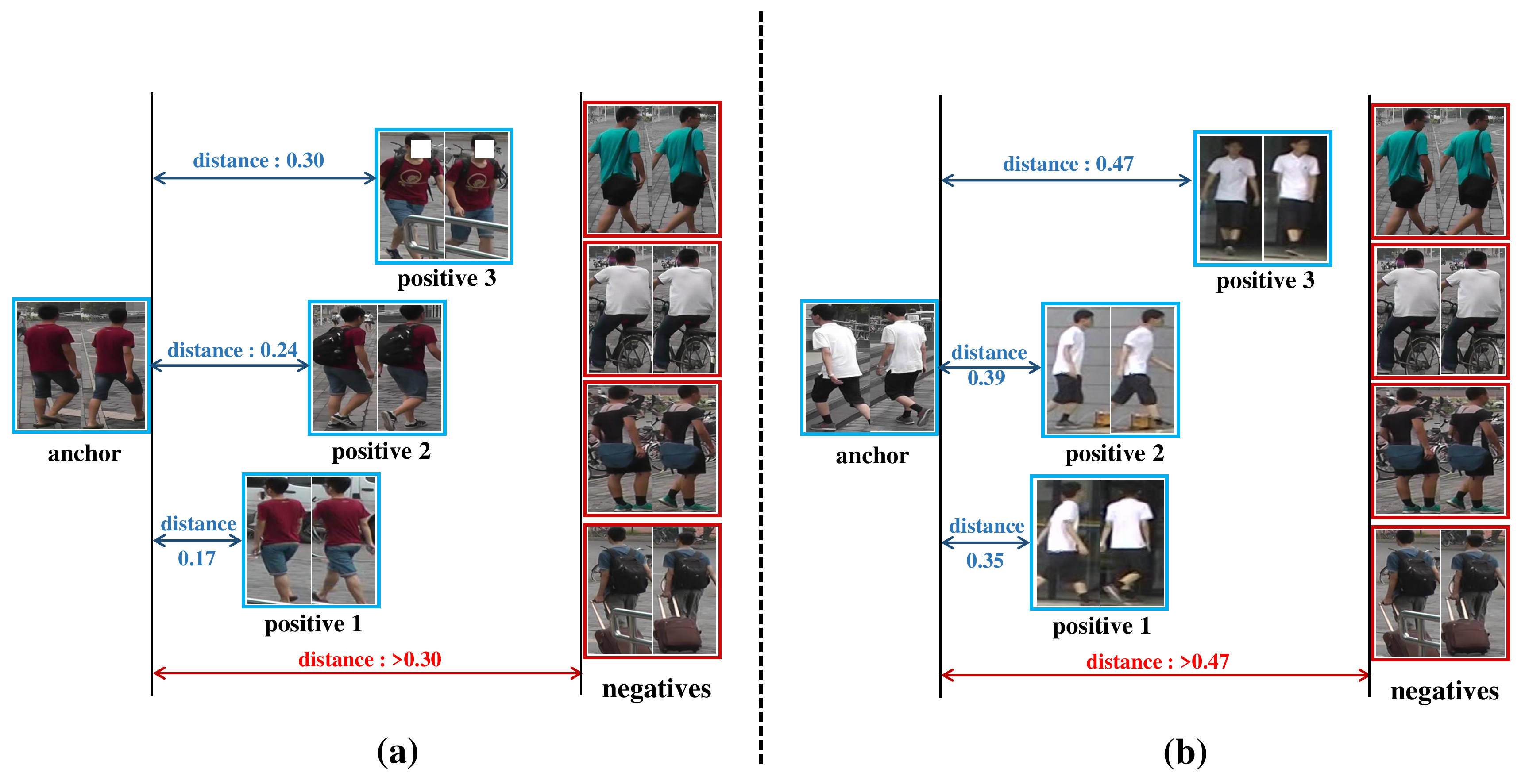}
	\vspace{-3mm}
	\caption{Normal batch-hard triplet loss can extremely shorten the distance between anchor and its corresponding positives and expand the distance to its negatives. But among different positives, there exists a high distance variance, the anchor tend to be much closer to the positive examples which has more similar pose or appearance with it.}

	\label{fig:triplet}
	\vspace{-5mm}
\end{figure}

In recent years, person re-identification (Re-ID) under video settings has drawn significant attention. In recent Re-ID studies, batch-hard triplet loss~\cite{hermans2017defense} is frequently used. This metric learning method can significantly narrow the distance between the anchor and its positives, and expand the margin between the anchor and its negatives in a mini-batch. But as shown in Figure~\ref{fig:triplet}, normal batch-hard triplet loss would cause the \emph{Distance Variance among Different Positives (DVDP) problem}, which makes the model less robust to intra-class variations such as pose and appearance difference. 

The attribute information has also been widely used for improving the performance of person re-identification. Many methods use the attribute to strengthen the correlation of image pairs or triplets~\cite{bmvc2012reid,tcsvt2015reid,icb2015reid,tpami2018reid,cvpr2018reid}, in which the distance between predicted attributes across identities is widely used. Some methods use attribute to help co-training the Re-ID models~\cite{lin2017improving,song2019two,tpami2018reid,cvpr2018reid,icpr2016reid,ling2019improving}. Recently, some feature aggregation strategies are adopted to make full use of the attribute information, as in~\cite{ding2019feature,han2018attribute,li2018diversity}. Although in the aforementioned methods introducing attribute to person Re-ID demonstrates good performance improvement, they are mainly practiced in an image-based way. Zhao et al.~\cite{zhao2019attribute} firstly apply attribute information to video-based person re-ID by introducing temporal attention learned from attribute recognition progress into the video-based Re-ID task via transfer learning. It should be pointed out that the frame-level temporal attention module in~\cite{zhao2019attribute} is actually trained from an image based RAP~\cite{li2016richly} pedestrian attribute dataset. However, the authors only use the attribute information to generate the temporal attentions and no spatial attention is considered. 

\begin{figure}[t]
	\centering
	\includegraphics[width=0.95\textwidth]{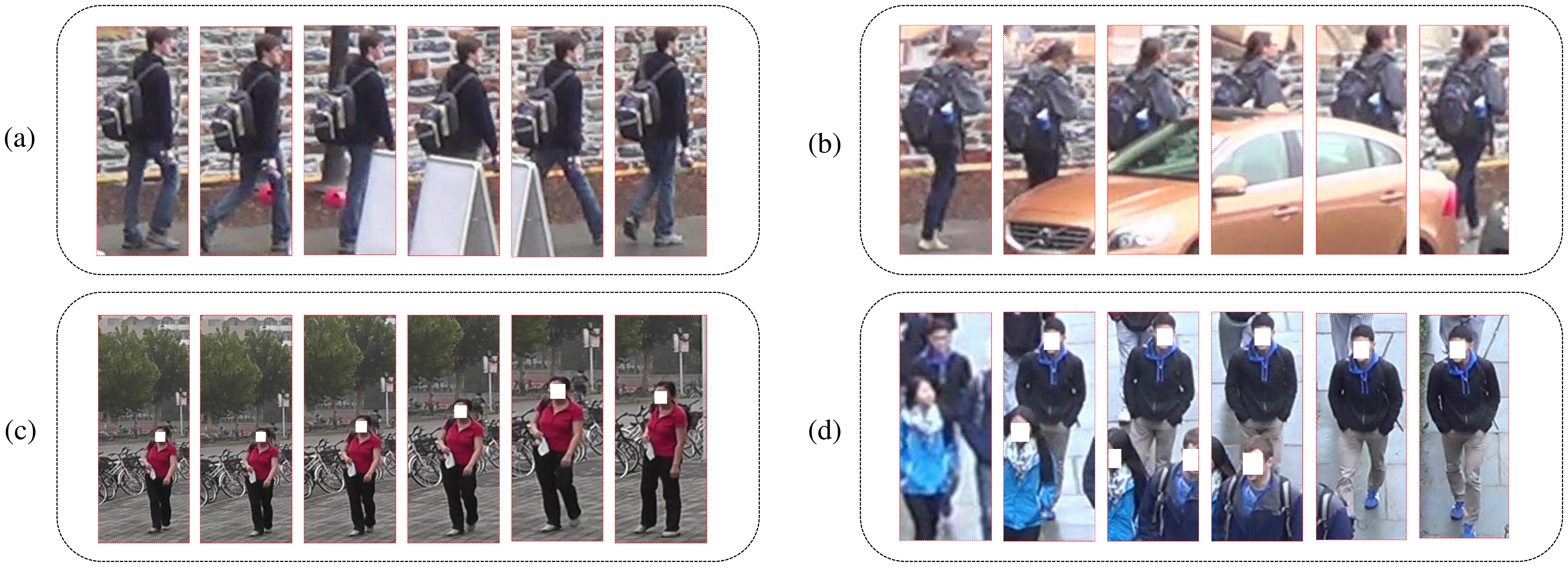}
	\vspace{-2mm}
	\caption{Performance of attribute recognition can be greatly influenced by temporal partial occlusion. Ruling out such negative effectives requires attention in both spatial and temporal axis. And the attribute-driven spatio-temporal attention can be also used for re-identifying people.}
	\label{fig:scores}
	\vspace{-5mm}
\end{figure}

As shown in Figure~\ref{fig:scores}(a) and Figure~\ref{fig:scores}(b), if the bottom part of a pedestrian is occluded, recognizing attribute such as \emph{bottom color} will be not applicable, and other quality problems in pedestrian videos such as background-dominated and multi-persons also affects the recognition of specific attributes, it shows that spatial cues are also very important in recognizing attribute. However, spatial attention are ignored in existing work of attribute-assisted video-based person Re-ID. Lack of annotation is the possible reason. Recently, attribute annotations for large video-based person Re-ID datasets become publicly available~\cite{chen2019temporal}, based on which real attribute-driven spatio-temporal attention can be learned. 

Based on above observations, a novel attribute-assisted approach for video-based person Re-ID is proposed in this paper. To address the DVDP problem, we propose the Attribute-aware Identity-hard Triplet Loss (AITL). By building triplets within positive samples of the same identity according to the attribute distance, the high DVDP can be considerably reduced and the Re-ID performance can be improved. To achieve a complete video-based person Re-ID approach, a multi-task framework with Attribute-driven Spatio-Temporal Attention (ASTA) mechanism is also proposed, in which attribute modeling, identity recognition and attribute embedding for DVDP reduction are integrated via an unified loss. The effectiveness of AITL and ASTA is well demonstrated by the experiments on MARS~\cite{zheng2016mars} and DukeMTMC-VID~\cite{wu2018exploit} dataset.

The main contributions in this paper are summarized as follows.
\begin{itemize}
	\item An attribute-aware metric learning method for the Distance Variance among Different Positives problem.
	\item A three-stream multi-task framework for attribute-assisted video-based person Re-ID using both identity-relevant and irrelevant attributes.
	\item An attribute-driven spatio-temporal attention mechanism for video-based person Re-ID.
\end{itemize}
The rest of this paper is organized as follows. In the next section we briefly review relevant research works. The attribute-aware identity-hard triplet loss is described in Section~\ref{sec:attribute-aware}. Then, Section~\ref{sec:model} introduces the proposed attribute-driven spatio-temporal attentive multi-task framework for video-based person Re-ID. Experimental results are presented in Section~\ref{sec:exp} and conclusion is drawn in Section~\ref{sec:summary}.

\section{Related Work}

\noindent\textbf{Video-based person Re-ID} Since person Re-ID is an application of video surveillance, video-based setting is a natural choice and closer to real-world scenario. Many challenging datasets have been created for video-based person Re-ID: iLIDS-VID~\cite{hamdoun2008person}, PRID2011~\cite{hirzer2011person}, MARS~\cite{zheng2016mars}, DukeMTMC-VID~\cite{wu2018exploit}, among these datasets, MARS and DukeMTMC-VID are frequently used recently, due to the large number in both track-lets and pedestrian identities. 

Early methods on video-based person Re-ID focus on handcrafting video representations and metric learning~\cite{wang2014person,gou2016person,liu2015spatio,you2016top,wang2016person}. Since the breakthrough of deep Convolutional Neural Network (CNN) and Recurrent Neural Networks (RNN), deep learning becomes the mainstream approach for video-based person Re-ID~\cite{li2014deepreid,ahmed2015improved,varior2016gated,wang2016joint,chen2017beyond,zhao2017spindle,zhang2018learning,liu2018video,revaud2015epicflow,zhang2017video,song2018mask,zhu2018novel,hou2019vrstc,subramaniam2019co}. McLaughlin et al.~\cite{mclaughlin2016recurrent} firstly proposed a baseline CNN-RNN model to extract features from the pedestrian track-lets. Liu et al.~\cite{liu2017quality}, Li et al. ~\cite{li2018diversity} and Song et al.~\cite{song2018region} estimated quality scores or high-quality regions for each frame to weaken the influence of noisy samples automatically. 

Siamese network has been adopted to video-based Re-ID recently~\cite{chung2017two,li2018multi}. Chen et al. ~\cite{chen2018video} proposed a competitive similarity aggregation scheme with short snippet-based representations to estimate the similarity between two pedestrian sequences. Gao and Nevatia~\cite{gao2018revisiting} make a thorough experimental survey on the temporal models in the video-based person Re-ID. Recently, the performance of the video-based Re-ID methods grow rapidly, Liu et al.~\cite{LiuSpatially} use non-local layers to apply the spatial and temporal attention into the training progress and get good performance on MARS and DukeMTMC-VID datasets. 


\noindent\textbf{Pedestrian Attribute Recognition} Similar to person Re-ID, early works on pedestrian attribute recognition are mainly image-based. Large dataset such as PETA~\cite{deng2014pedestrian}, RAP~\cite{li2016richly}, P-100K~\cite{liu2017hydraplus} have been created. Traditional handcrafting methods as well as end-to-end deep learning methods are all adopted~\cite{lin2017improving,bmvc2012reid,deng2014pedestrian,tcsvt2015reid,iccvw2013attr,iccvw2015attr,acpr2015attr,liu2017hydraplus,ijcai2018attr,icme2018attr}.

Chen et al.~\cite{chen2019temporal} first tackle the pedestrian attribute recognition problem by video-based settings and provide attribute annotations for MARS DukeMTMC-VID datasets. They propose a temporal attention based method and demonstrate the superiority of using video in recognizing attributes. 


\noindent\textbf{Attribute-assisted Person Re-ID} Attribute such as gender, age and clothing can be viewed as some kind of ``soft biometrics''.Therefore, they can provide additional information and have been introduced into person Re-ID. Early methods use the attributes to strengthen the correlation between image pairs or triplets~\cite{bmvc2012reid,tcsvt2015reid,icb2015reid,tpami2018reid,cvpr2018reid}. Some methods use attributes as a strong supervision to help co-training~\cite{lin2017improving,song2019two,tpami2018reid,cvpr2018reid,icpr2016reid,ling2019improving}. In some latest researches, feature aggregation strategy is adopted to make full use of the attribute information~\cite{ding2019feature,han2018attribute}. 

Song et al.\cite{song2019two} firstly apply the attribute information in video-based person re-ID problem. However, this work is rather limited for the reason of lack of data. Zhao et al.\cite{zhao2019attribute} apply attribute to video-based person re-ID by learning temporal attention from the external image-based RAP dataset. Although they clearly show the effectiveness of cooperating with attributes, their attention model is limited to temporal dimension for the reason of lack of attribute ground truth on the video-based person Re-ID datasets.

\noindent\textbf{Metric Learning for Person Re-ID} Except for the batch-hard triplet loss~\cite{hermans2017defense}, some other metric learning methods are also applied to handle the person Re-ID task, such as Quadruplet loss~\cite{chen2017beyond} and Margin sample mining loss~\cite{xiao2017margin}, these metric learning method also focus on improving the Re-ID performance by adjusting the distance between positive and negative pairs. Among all these methods, batch-hard triplet loss is most commonly used in recent Re-ID researches due to the good generalization ability.

\section{Attribute-aware Identity-hard Triplet Loss}
\label{sec:attribute-aware}

As introduced in Section~\ref{sec:introduction} and Figure~\ref{fig:triplet}, batch hard triplet loss is really effective in narrowing the distance between anchor and its positives, and also in expanding the distance between anchor with its negatives. But it ignores the distance relation among positives. As shown in Figure~\ref{fig:triplet}, normal batch-hard triplet loss would cause a problem that when the training process tends to be stable, the anchor-to-negative-distances are usually much larger than the anchor-to-positives-distance. Even the overall triplet loss tends to be zero, there still exists a distance variance between anchor and its different positives. The feature of anchor would be much closer to the positives which is more similar in appearance. This would lead the model to pay more attention to the direct appearance of the pedestrian, and be less robust to pose and appearance variances between different positives. Obviously, it is harmful to person Re-ID.

So in this paper, we designed an attribute-aware identity-hard triplet loss to solve this problem. We use the attribute prediction vectors to measure the appearance similarity among different positive pairs, and then fix the Re-ID feature distance gap (see Figure~\ref{fig:dvdp}) between the anchor to the most similar positive and the anchor to the most different positive. 

\begin{figure*}[t]
	\centering
	\includegraphics[width=0.95\textwidth]{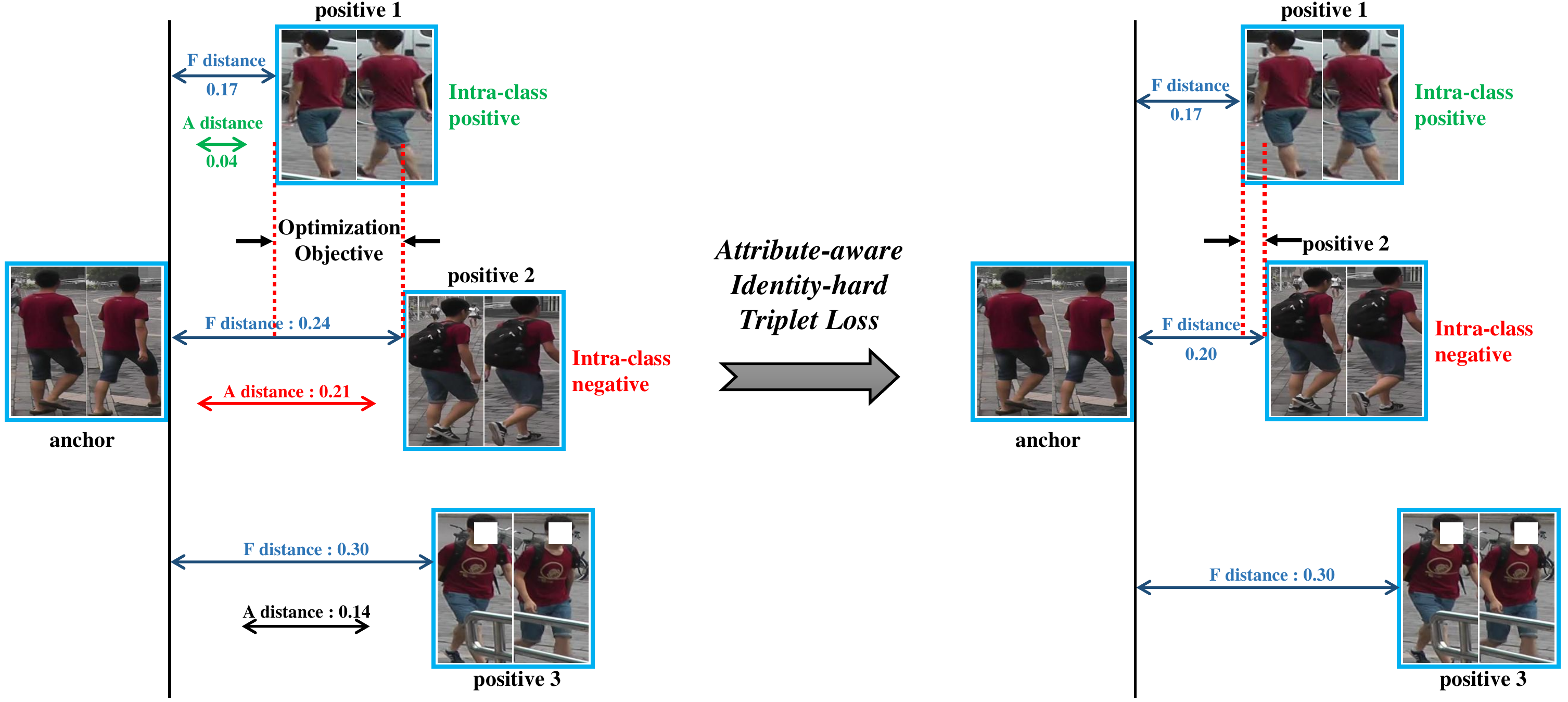}
	\vspace{-2mm}
	\caption{Brief illustration of the attribute-aware triplet loss. The ``A distance'' denotes the distance between attribute prediction vectors of different pedestrian videos, while ``F distance'' means the distance of Re-ID features. Due to the DVDP problem, there is a big difference among the anchor-to-positive distances (left). We find that the attribute distance is negative correlated to DVDP. By building additional triplets within the identity positives according to the attribute distance and minimizing the negative loss, the identity-level DVDP can be reduced.}
	\label{fig:triplet2}
	\vspace{-5mm}
\end{figure*}

\textbf{Intra-Class Triplet Loss} In this paper, we introduce the attribute-aware triplet loss into the positives themselves to solve the DVDP problem. As shown in Figure~\ref{fig:triplet2}, the attribute prediction vectors generated by the multi-task model, which is described in Section~\ref{sec:model}, can be used to measure the appearance similarity between anchor and its corresponding positives.

Based on the attribute distance, \emph{intra-class negative} and \emph{intra-class positive} can be picked out from the \emph{identity positives}. As shown in Figure~\ref{fig:triplet2}, their distances to the anchor are usually different. In other words, at the identity-level, the intra-class variation a.k.a the DVDP is highly correlated to the attribute difference. And the high DVDP pairs can be automatically discovered by the attribute.
Once the \emph{intra-class negative} and \emph{intra-class positive} are selected, triplet based optimization can be performed to narrow the distance. 

\textbf{Attribute-aware} As mentioned above, we use the attribute distance rather than Re-ID feature distance to select \emph{attribute-aware intra-class negative} and \emph{attribute-aware intra-class positive}. The reason is that the Re-ID features of the pedestrian videos is a high-level semantics while the attribute prediction result is low-level. Low-level semantics is naturally more suitable to be used to compare the appearance similarity between different pedestrian videos and be more robust to noisy and hard samples. The advantages of attribute distance in selecting \emph{intra-class negative} and \emph{intra-class positive} would also be illustrated through ablation study in Section~\ref{sec:ablation}. 

\textbf{Identity Hard} To form a mini-batch, the attribute-aware triplet loss basically follows the sampling rule proposed by Hermans et al.~\cite{hermans2017defense}. The core idea is to form batches by randomly sampling $P$ classes (people), and then randomly sampling $K(K >= 3)$ videos of each class, thus resulting in a batch of $PK$ videos. The only difference between the AITL and Hermans et al.~\cite{hermans2017defense} is that the attribute-aware triplet loss select the \emph{intra-class negative} and \emph{intra-class positive} only from the samples sharing the same identity label with the anchor, so theoretically its \emph{Identity Hard}. 

For each anchor $a$, we can select the \emph{intra-class positive} and the \emph{intra-class negative} within the samples sharing the same identity with $a$ in the batch when forming the attribute-aware triplets. The Distance Variance among Different Positives(DVDP) on this single batch can be represented by
\begin{small}
	\begin{equation}
	\begin{array}{l}
	\mathcal{DVDP}(\theta;X)=\overbrace{\sum_{i=1}^{P}\sum_{a=1}^{K}}^{all\ anchors}[\overbrace{FD(f_{\theta}(x_a^i), f_{\theta}(x_{FN(i, a)}^i))}^{Intra-class\ negative} - \underbrace{FD(f_{\theta}(x_a^i), f_{\theta}(x_{FP(i, a)}^i))}_{Intra-class\ positive}]_+.
	\end{array}
	\label{equ:1}
	\end{equation}
\end{small}

And the Attribute-aware Identity-hard Triplet Loss which is used to reduce DVDP can be written as
\begin{small}
	\begin{equation}
	\begin{array}{l}
	\mathcal{L}_{AITL}(\theta; X)=\overbrace{\sum_{i=1}^{P}\sum_{a=1}^{K}}^{all\ anchors}[\overbrace{FD(f_{\theta}(x_a^i), f_{\theta}(x_{AN(i, a)}^i))}^{attribute\ negative} - \underbrace{FD(f_{\theta}(x_a^i), f_{\theta}(x_{AP(i, a)}^i))}_{attribute\ positive}]_+.
	\end{array}
	\label{equ:1}
	\end{equation}
\end{small}

This is defined for a mini-batch $X$ and where data point $x_{j}^{i}$ corresponds to the $j$-th video of the $i$-th person in the batch. In Equation~\eqref{equ:1}, $\theta$ means the parameters of the Re-ID feature function $f$, $FD$ denote the Re-ID feature distance between different samples, $FN$ and $FP$ means the index of the \emph{intra-class negative} and \emph{intra-class positive} in feature distance, $AN$ and $AP$ means the index of the \emph{attribute-aware intra-class negative} and \emph{attribute-aware intra-class positive} for the anchor, for each anchor which indexed $i,a$, the $FN$ and $FP$, $AN$ and $AP$ can be represented by
\begin{equation}
\begin{array}{l}
FN(\gamma; i, a) = \underset{\underset{j\neq a}{j=1,...,K}}{argmax}\ FD(g_{\gamma}(x_a^i), g_{\gamma}(x_j^i)) \\
FP(\gamma; i, a) = \underset{\underset{j\neq a}{j=1,...,K}}{argmin}\ FD(g_{\gamma}(x_a^i), g_{\gamma}(x_j^i)).\\
AN(\gamma; i, a) = \underset{\underset{j\neq a}{j=1,...,K}}{argmax}\ AD(g_{\gamma}(x_a^i), g_{\gamma}(x_j^i)) \\
AP(\gamma; i, a) = \underset{\underset{j\neq a}{j=1,...,K}}{argmin}\ AD(g_{\gamma}(x_a^i), g_{\gamma}(x_j^i)).\\
\end{array}
\label{equ:2}
\vspace{-5mm}
\end{equation}

In Equation~\eqref{equ:2}, $\gamma$ means the parameters of the attribute recognition model $g$, and $AD$ means the distance on the attribute prediction vector between different pedestrian videos. For the distance function $AD$ and $FD$, we use the cosine distance as the metric.

By using the attribute prediction vector to calculate the appearance similarity between anchor and different positives, we could easily get the \emph{intra-class negative} and \emph{intra-class positive} for the anchor by comparing the attribute distance. And by narrowing the Re-ID feature distance gap between the \emph{intra-class negative} and \emph{intra-class positive}, the Distance Variance among Different Positives would be considerably reduced. Consequently, the resulted model is more robust to variations like pose difference. 

\begin{figure*}[t]
	\centering
	\includegraphics[width=0.99\textwidth]{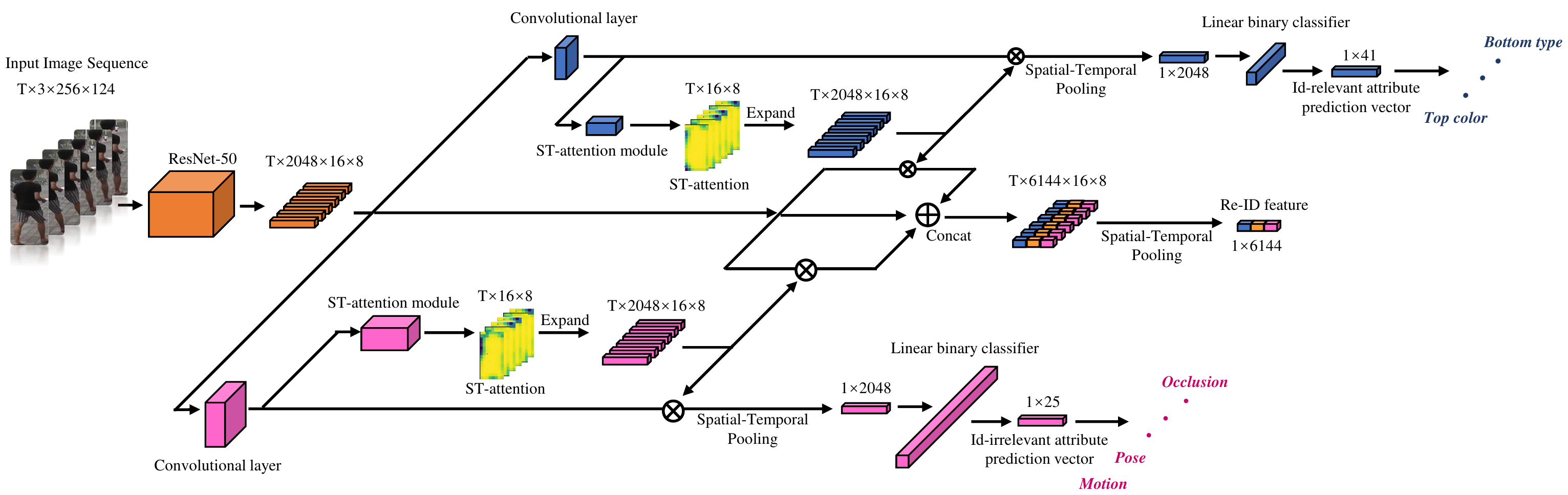}
	\vspace{-2mm}
	\caption{Architecture of our multi-task framework. It is mainly composed of three streams: The Re-ID backbone, ID-irrelevant attribute recognition stream for ID-irrelevant Attribute Recognition, as well as the ID-relevant attribute recognition stream for recognizing ID-relevant attributes.}
	\vspace{-5mm}
	\label{fig:network}
\end{figure*}

\section{Multi-task Framework based on Attribute-driven Spatio-Temporal Attention}
\label{sec:model}

Video data can provide more information than single image. However, as shown in Figure~\ref{fig:scores}, due to inaccurate detector and tracker, noisy frames are inevitably contained. Therefore, as well as picking out discriminative frames, screening off disturbing frames and regions is also important for video-based Re-ID. Learning attentions from middle-level attribute is a better way for feature refinement.

The appearance of a pedestrian sequence is influenced by two kinds of factors. The internal factor such as clothing characteristics is directly relevant to identity, therefore should be emphasized. The external factors such as pose angle and occlusion are harmful to recognition, to improve the performance, their impact should be reduced. Fortunately, the recently released dataset~\cite{chen2019temporal} provides annotations for both kinds of factors, which makes it possible to learn a comprehensive spatio-temporal attentions from pedestrian attribute. 

Based on above observations, we propose a multi-task framework for attribute-enhanced video-based Re-ID. It consists of three streams, which correspond to identity-relevant attribute, identity itself, and identity-irrelevant attribute. As can be seen from Figure~\ref{fig:network}, to achieve final re-identification, the three channels are fused together via transferring spatio-temporal attentions from attribute.

\subsection{overview}

As illustrated in Figure~\ref{fig:network}, the multi-task framework consists of three streams. Before stream splitting, a backbone ResNet-50 is used to extract the basic frame feature. The identify-irrelevant  (ID-irrelevant) attribute recognition stream is used to recognize attribute like motion, pose and occlusion. It takes the frame features as input, then feeds them into an additional size-preserving convolutional layer. After that the feature is further processed by a spatio-temoral attention module, and get the attribute-driven spatio-temoral attention whose size is $T\times16\times8$, where $T$ is the frame number. The attention vector can be expanded into original feature size for later point-wise multiplication. Finally, a spatio-temporal pooling operation is performed to encode the feature to $1\times2048$. It is achieved by performing average pooling in both spatial and temporal axis respectively. Based on the final feature, attribute estimation can be achieved by using a linear classifier. 

The structure of identity-relevant (ID-relevant) attribute module is the same except for the size of output attribute prediction. ID-relevant attribute recognition module mainly focus on the attribute related to the pedestrian identity like the color of clothes, gender, etc, while the ID-irrelevant attribute recognition module pay attention to motion, pose, and noises like occlusion. The output attribute prediction vectors generated from these two attribute recognition streams could be used for attribute-aware identity-hard triplet loss, which has been discussed in Section~\ref{sec:attribute-aware}.

Since each attribute stream can generate a spatial-temporal attention vector, which contains rich spatio-temporal information learned in attribute recognition progress, the features from the Re-ID are encoded by the two attribute-driven spatial-temporal attention vectors respectively. The two attention enhanced features are concatenated with the original one. By applying a spatio-temporal pooling operation, the overall feature is finally encoded to $1\times6144$. This final feature not only keeps the information from original frames, but also combines the spatial-temporal clues learned from attributes.
\begin{figure}[t]
	\centering
	\includegraphics[width=0.7\textwidth]{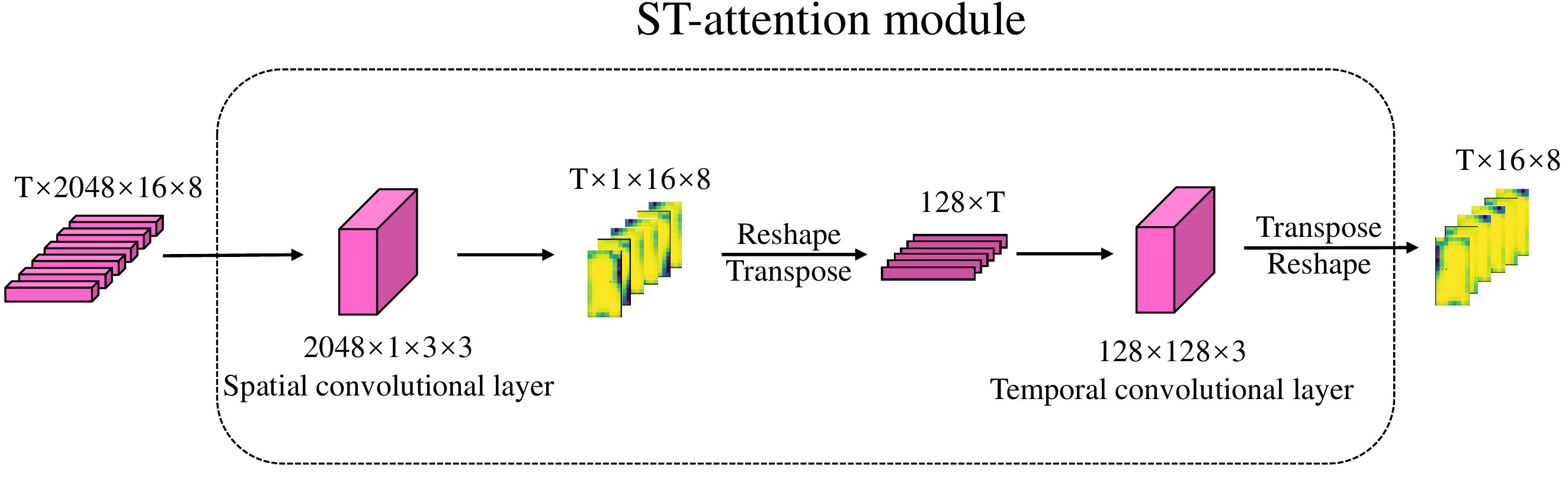}
	\vspace{-2mm}
	\caption{Detailed structure of our spatio-temporal attention module.}
	\label{fig:attention}
	\vspace{-5mm}
\end{figure}
\subsection{Spatio-Temporal Attention Module}

To generate spatio-temporal attention vector from the frame features, we designed a light-weighted yet effective spatial-temporal attention module. As shown in Figure~\ref{fig:attention}, the spatio-temporal attention module takes the frame features as input. Firstly, the channel dimension of the frame feature is reduced to one by a two dimensional convolutional layer and output the spatial attention vector. Then after reshape and transpose operation, the attention vector is converted to $128\times T$. In the next, the attention vector would be processed by a one dimensional temporal convolutional layer, the temporal attention layer is a 1-d convolutional layer that the number of input channel and output channel is 128, kernel size is 3, padding is 2, stride is 1. It takes the 128 * T (T is the temporal axis) spatial attention vector as input and generate the ST-attention vector which has the same size with input by conduct 1-d convolution operations on the temporal axis. Finally, the attention vector would be turned into original shape, and after the sigmoid activation, this spatio-temporal vector would be used to encode the attribute recognition feature as well as the Re-ID feature.

\subsection{Unified Loss}

Since the multi-task model combines the attribute recognition task and Re-ID task, we unified several loss functions in the training process. We use \emph{Binary Cross Entropy Loss} $\mathcal{L}_{BCE}$ to train the attribute recognition module. For Re-ID task, besides the normal loss function combination of batch-hard triplet loss $\mathcal{L}_{tri}$ and softmax loss $\mathcal{L}_{softmax}$, we introduced the attribute-aware identity-hard triplet loss $\mathcal{L}_{AItri}$ which has been discussed in Section~\ref{sec:attribute-aware} to solve the DVDP problem. So the final loss function of the multi-task model in the training progress can be written as
\begin{equation}
\mathcal{L} = \overbrace{\mathcal{L}_{BCE}}^{attributes} + \underbrace{\mathcal{L}_{tri} + \mathcal{L}_{softmax} + \mathcal{L}_{AITL}}_{Re-ID}.
\end{equation}

As the final Re-ID feature is the result of concatenation, to strike a balance between attribute and identity, we use cosine distance in the two triplet loss, which is equivalent to squared euclidean distance if the features is normalized.

\begin{table}[t]
	\caption{Ablation study on person Re-ID task on MARS(\%).}
	\centering
	\begin{tabular}{p{3.7cm}|c|c|c|c}
		\toprule
		\multirow{2}*{Model} & \multicolumn{4}{c}{MARS}\\
		\cline{2-5}
		{}&R1&R5&R10&mAP\\
		\toprule
		\bottomrule
		Baseline.&84.9&95.2&96.6&79.4 \\
		\cline{1-5}
		Baseline. + ASTA&86.6&95.8&97.2&82.4 \\
		\bottomrule
		\toprule
		Baseline. + ITL&86.7&96.0&97.4&83.3 \\
		\cline{1-5}
		Baseline. + AITL&87.4&96.2&97.6&83.5 \\
		\bottomrule
		\toprule
		Baseline. + ASTA + AITL&\textbf{88.2}&\textbf{96.5}&\textbf{97.8}&\textbf{84.4} \\
		\bottomrule
	\end{tabular}
	\label{table:ablationstudy}
\end{table}
\section{Experiments}
\label{sec:exp}
We evaluate our method on two large video datasets for person Re-ID, i.e. the MARS~\cite{zheng2016mars}, and DukeMTMC-VID~\cite{wu2018exploit} respectively. MARS consists of 1,261 people captured from six cameras, in which 625 are used for training and the rest are for test. The DukeMTMC-VID is a subset of the DukeMTMC~\cite{ristani2016performance}, which consists of 702 training subjects, 702 test subjects and 408 distractors. The track-lets in both dataset are automated detected and tracked. The attribute annotation for MARS and DukeMTMC-VID is provided by Chen et al.~\cite{chen2019temporal}.

In the experiments, all the detected pedestrian are resized to $128\times256$. We set the video clips size $T = 4$ in the training progress, which just follows the best setting of the baseline model in Gao and Nevatia~\cite{gao2018revisiting}. All the models in the experiments are implemented in Pytorch, we choose Adam as the optimizer, and the learning rate is set to 0.0003. The source codes of this paper would be published in the future.
\subsection{Ablation study}
To verify the effectiveness of our multi-task model and the Attribute-aware Identity-hard Triplet Loss, we trained several models on the MARS dataset: the baseline temporal pooling model by Gao and Nevatia~\cite{gao2018revisiting}; baseline with the Attribute-driven Spatio-Temporal Attention (ASTA); baseline with Identity-hard Triplet Loss (ITL) which selects the intra-class positive and negative by directly comparing the feature distance rather than attribute distance; baseline with Attribute-aware Identity-hard Triplet Loss (AITL); the proposed multi-task model which combines the Attribute spatio-temporal Attention mechanism as well as the Attribute-aware Identity-hard Triplet Loss.
\begin{figure}[t]
	\centering
	
	\includegraphics[width=0.95\textwidth]{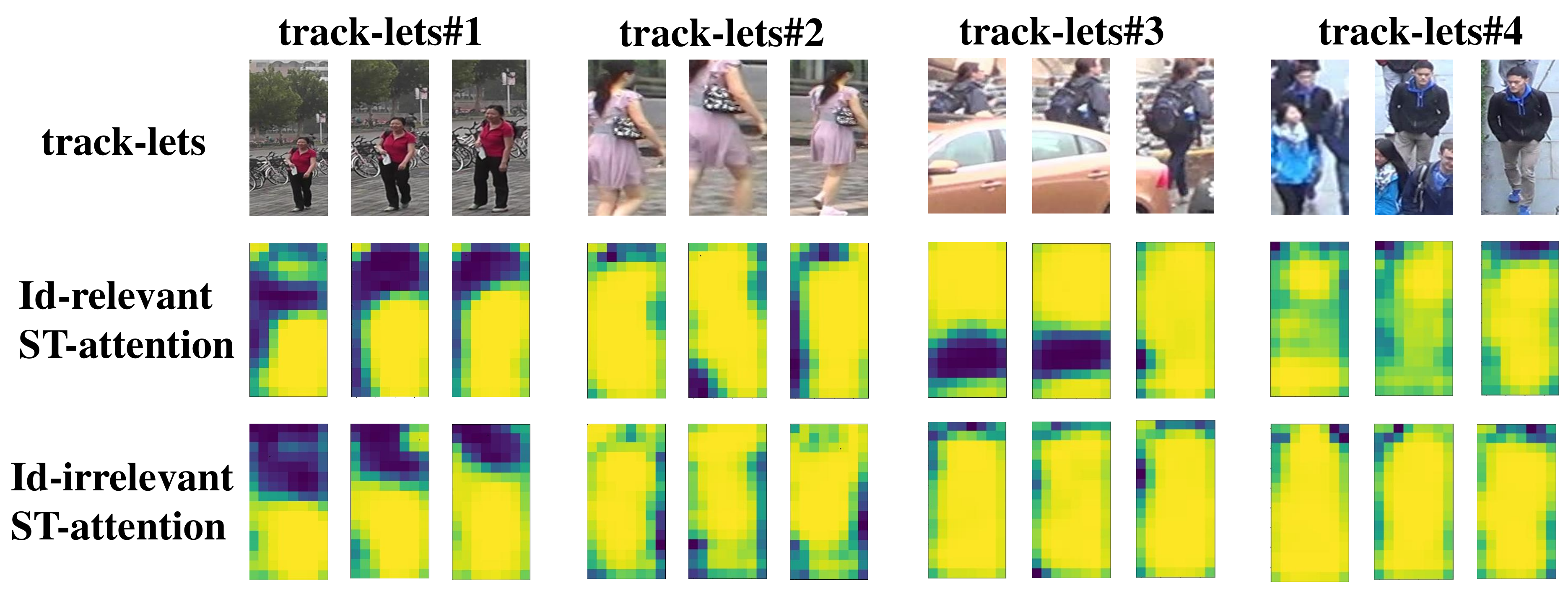}
	\vspace{-5mm}
	\caption{Visualization of ST-attention on low-quality track-lets.}
	
	\label{fig:visualization}
	\vspace{-5mm}
\end{figure}

\begin{table}[b]
	\vspace{-4mm}
	\caption{Results of different stream combinations on MARS(\%).}
	\centering
	\begin{tabular}{ccc|c|c}
		\toprule
		\multicolumn{3}{c}{Stream combination}&\multicolumn{2}{c}{Metrics}\\
		\cline{1-5}
		Re-ID&ID-relevant&ID-irrelevant&R1&mAP\\
		\toprule
		\bottomrule
		\checkmark&&&84.9&79.4\\
		&\checkmark&&87.1&83.7\\
		&&\checkmark&86.6&81.5\\
		\checkmark&\checkmark&&87.9&84.1\\
		\checkmark&&\checkmark&86.7&82.8\\
		&\checkmark&\checkmark&88.0&84.2\\
		\bottomrule
		\checkmark&\checkmark&\checkmark&\textbf{88.2}&\textbf{84.4}\\
		\bottomrule
	\end{tabular}
	
	\label{table:ablationstudy}
\end{table}
As shown in Table~\ref{table:ablationstudy}, by integrating the spatio-temporal attention learned in the attribute recognition progress to the person Re-ID task, the performance of re-identification can be consistently improved, which implies that the attribute recognition streams could find the discriminative spatial and temporal clues out from the pedestrian image sequence, and the discriminativeness is not only good for recognizing attribute but also helpful to re-identification. 

Without attention, solely using the Identity-hard Triplet Loss gains more consistent performance improvements. It's obvious that AITL outperforms ITL in every metric. Because directly using the feature distance to determine the intra-class positive and negative in a triplet is vulnerable to noisy and difficult samples. Introducing attribute distance could make the model more robust. Combining ASTA and AITL, the proposed multi-task model reaches the best performance, which proves the effectiveness of the strategies we proposed.

Different attribute streams have different effects on the improvement of Re-ID performance. As shown in Figure~\ref{fig:visualization}, for the same track-lets, the id-relevant and id-irrelevant attribute streams have different spatio-temporal concentrations, but they consistently focus on the human body. In other words, although the attributes are irrelevant to specific person, they are still closely relevant to general human body. Consequently, the learned attention can filter out background elements and occlusions. Recognition results (see Table~\ref{table:ablationstudy}) also show that both ST-attentions are beneficial. 

To the feature length issue, as can be seen from Table~\ref{table:ablationstudy}, even with the same length, the model trained with attribute is still much better than baseline, which implies that the improvement are mainly derived from attribute rather than the expansion of feature length.

\begin{figure}[t]
	\centering
	\begin{tabular}{c}
		\includegraphics[width=0.46\textwidth]{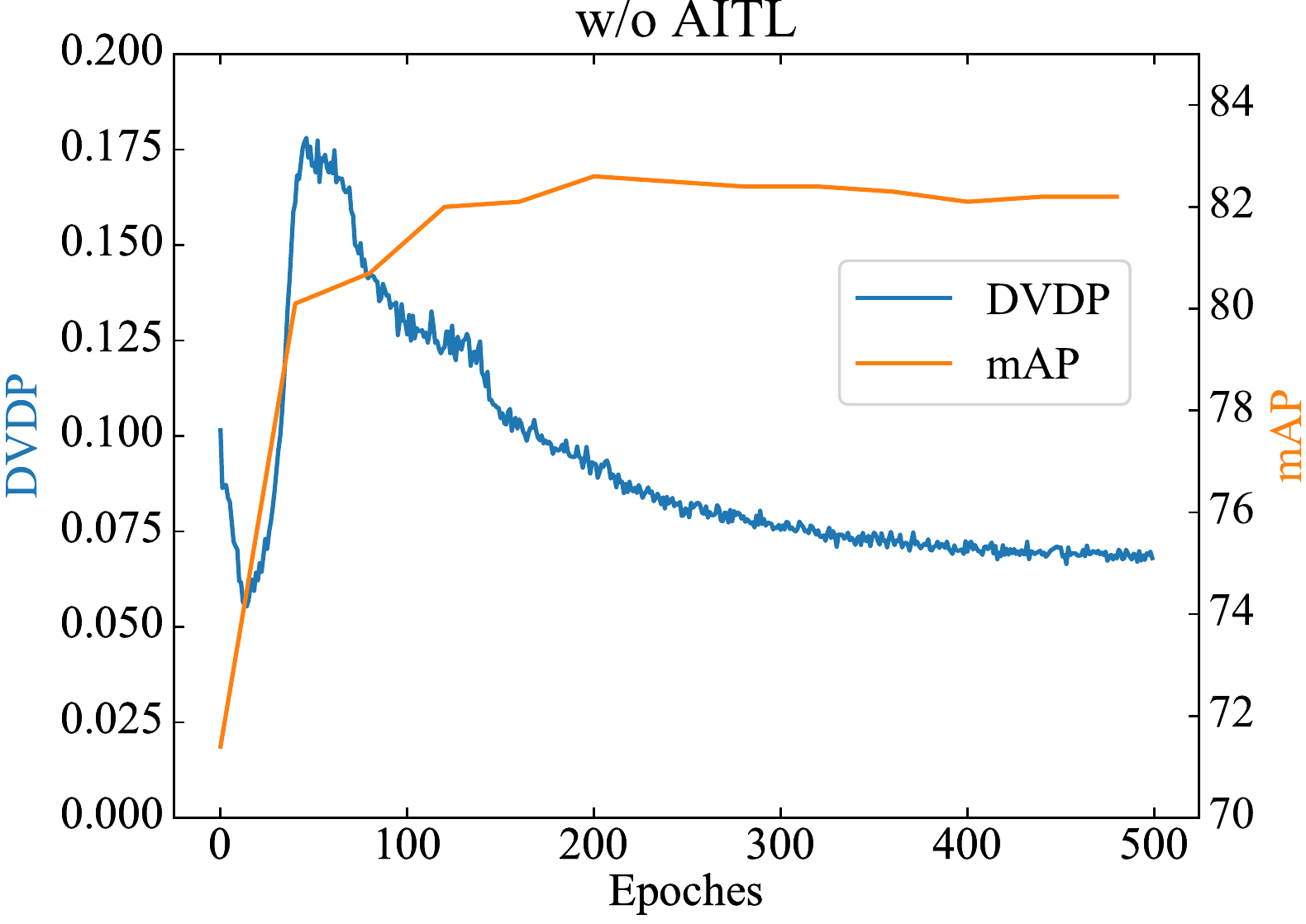}
		\includegraphics[width=0.46\textwidth]{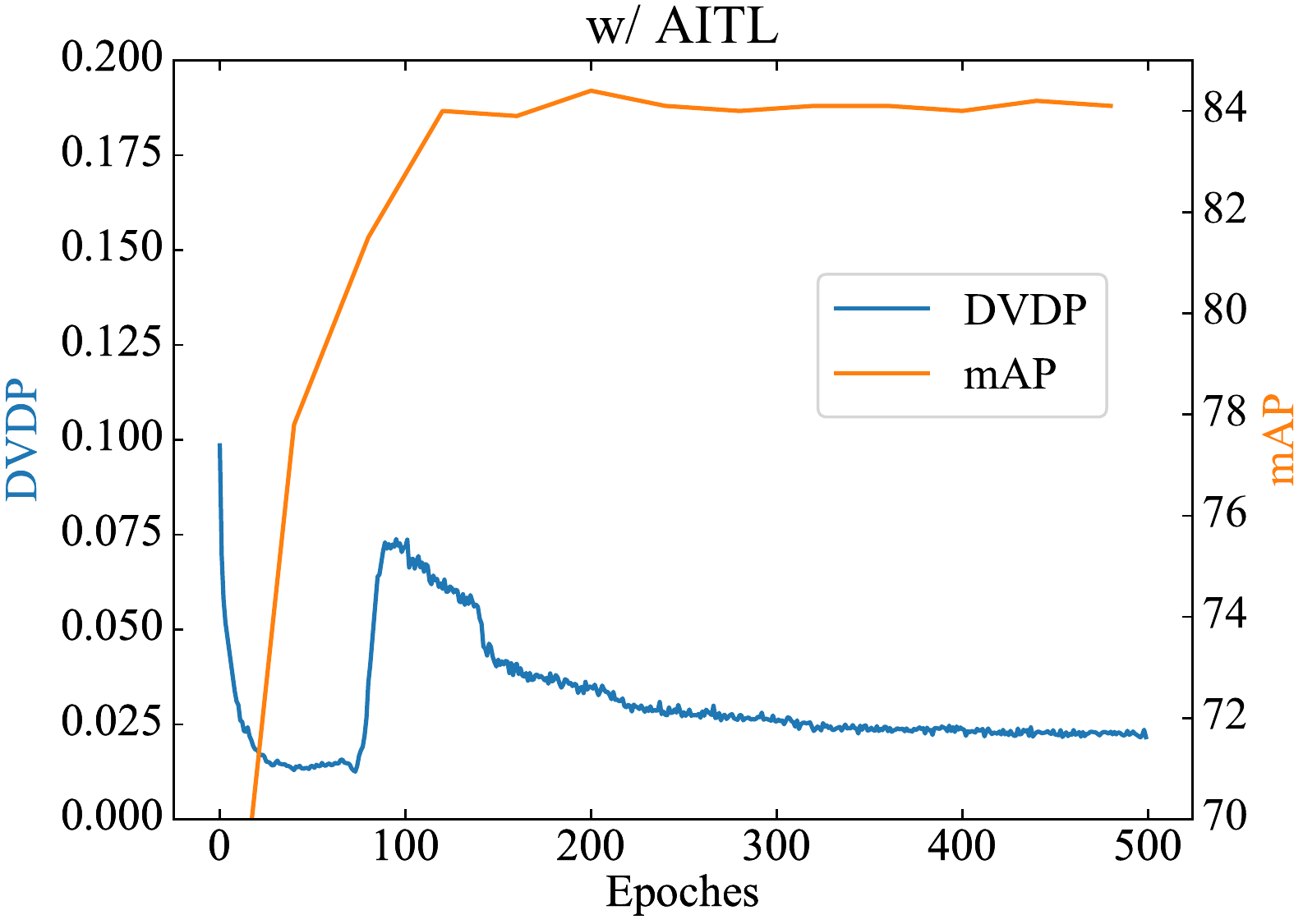}
	\end{tabular}
	\caption{Variation curve of DVDP and mAP in the training progress with or without the Attribute-aware Identity-hard Triplet Loss (AITL) supervision.}
	\vspace{-5mm}
	\label{fig:dvdp}
\end{figure}

\subsection{Influence of the Attribute-aware Identity-hard Triplet Loss}
\label{sec:ablation}

Although the ablation study shows the effectiveness of the Attribute-aware Identity-hard Triplet Loss in improving the Re-ID performance, it's still not intuitive enough if the AITL can really ease the DVDP problem. To illustrate the effectiveness of AITL in reducing the Distance Variance among Different Positives, we record the average DVDP in each epoch in the training progress by computing the average distance between the \emph{intra-class negative} and \emph{intra-class positive} for all anchors in each batch, as well as the mAP in every validation process. By observing the variation of this two variables, we can get a more intuitive illustration on the performance of AITL in reducing the DVDP and in improving the Re-ID performance.

As shown in Figure~\ref{fig:dvdp}, without AITL supervision, the DVDP would be stacked at about 0.075 as the training progress becomes stable, while with the AITL supervision, the DVDP quickly decreases. The mAP metric can also benefit from this decreasing in DVDP and reach a higher position than without AITL. This example clearly shows the effectiveness of the AITL in reducing the Distance Variance among Different Positives. 

In both two curves the DVDP drops dramatically at the beginning of training process, and then quickly increases. This phenomenon may be caused by the normal batch hard triplet loss which is often considered an unstable process in the beginning of training.

\begin{table}[t]
	\caption{Performance comparison on MARS (\%).}
	\centering
	\begin{tabular}{p{1.9cm}|c|c|c|c|c|c|c|c|c}
		\toprule
		\multirow{2}*{Model} & \multirow{2}*{Source} & \multicolumn{4}{c}{MARS} & \multicolumn{4}{c}{DukeMTMC-VID}\\
		\cline{3-10}
		{}&{}&R1&R5&R20&mAP&R1&R5&R20&mAP\\
		\toprule
		\bottomrule
		Zheng~\cite{zheng2016mars}&ECCV16&65.0&81.1&88.9&46.6&--&--&--&-- \\
		Hermans~\cite{hermans2017defense}&arXiv17&79.8&91.4&--&67.7&--&--&--&-- \\
		Xiao~\cite{xiao2017margin}&arXiv17&83.0&92.6&--&72.0&--&--&--&-- \\
		Liu~\cite{LiuSpatially}&BMVC19&\textbf{90.0}&--&--&82.8&\textbf{96.3}&--&--&94.9\\
		Zhao~\cite{zhao2019attribute}&CVPR19&87.0&95.4&\textbf{98.7}&78.2&--&--&--&-- \\
		Fu~\cite{fu2019sta}&AAAI19&86.3&95.7&98.1&80.8&96.2&99.3&99.6&94.9 \\
		Chen~\cite{chen2018video}&CVPR18&86.3&94.7&98.2&76.1&--&--&--&-- \\
		Hou~\cite{hou2019vrstc}&CVPR19&88.5&\textbf{96.5}&97.4&82.3&95.0&99.1&--&93.5 \\ 
		Li~\cite{li2019global}&ICCV19&87.0&95.8&98.2&78.59&\textbf{96.3}&99.3&99.7&93.7 \\ 
		\bottomrule
		ours&--&88.2&\textbf{96.5}&98.4&\textbf{84.4}&95.4&\textbf{99.6}&\textbf{99.9}&\textbf{95.3}  \\
		\bottomrule
	\end{tabular}
	\label{table:marsstoa}
\end{table}


\subsection{Comparison with State-of-the-art Methods}

We compared our method (w/ ASTA and AITL) with some state-of-the-art video-based person Re-ID methods on MARS and DukeMTMC-VID: 1) \textbf{Image-based method}: such as Resnet-50 baseline~\cite{zheng2016mars}; 2) \textbf{Triplet Loss}: Hermans et al.~\cite{hermans2017defense}, Margin sample mining loss proposed by Xiao et al.\cite{xiao2017margin}; 3) \textbf{Spatial or temporal attention method}: Temporal attention model by non-local layers~\cite{LiuSpatially}, Attribute-driven temporal aggregation model~\cite{zhao2019attribute}, Spatio-Temporal Attention Network~\cite{fu2019sta}; 4) \textbf{Other methods}, Co-attentive Snippet Embedding (CASE)~\cite{chen2018video}, VRSTC~\cite{hou2019vrstc}, Global-Local Temporal Representations~\cite{li2019global}.

The comparisons on MARS are shown in Table~\ref{table:marsstoa}. Compared with the image-based baseline method, identity-driven spatio-temporal attention models or the model trained with triplet loss, the proposed method can outperform them in all metrics. While compared with some latest methods with complex architectures, such as the models by Liu et al.~\cite{LiuSpatially}, which use non-local layers to enrich the local image feature with global sequence information by generating attention masks according to features of different frames and different spatial locations, and the VRSTC model~\cite{chen2018video} which use GAN-like model to recover the appearance of occluded parts, our model can still achieve the best performance in mAP and Rank-5 recognition rate. To our knowledge, the work by Zhao et al.~\cite{zhao2019attribute} is the best approach in attribute-assisted video-based person Re-ID. Our method can consistently outperform it except for Rank-20 recognition rate.

Several state-of-the-art video-based Re-ID methods are also compared on the DukeMTMC-VID. The experimental results are shown also in Table~\ref{table:marsstoa}. Our method can clearly outperform all the aforementioned approaches except for the Rank-1 metric. The mAP margin on DukeMTMC-VID is not as large as it on MARS. Because in training the AITL, it's better to select $K\geq3$ videos from each identity. However, most people in the training set of DukeMTMC-VID cannot meet this requirement. That's limits the performance of AITL.

In conclusion, by comparing our multi-task model with state-of-the-art methods, we can prove that the proposed attribute-aware spatio-temporal attention and identity-hard triplet loss are very effective. It should be pointed out that the baseline model used in our method is rather simple. The performance could be further improved by introducing better baseline models.

\subsection{Computation-performance}
\begin{figure*}[t]
	\centering
	\includegraphics[width=0.8\textwidth]{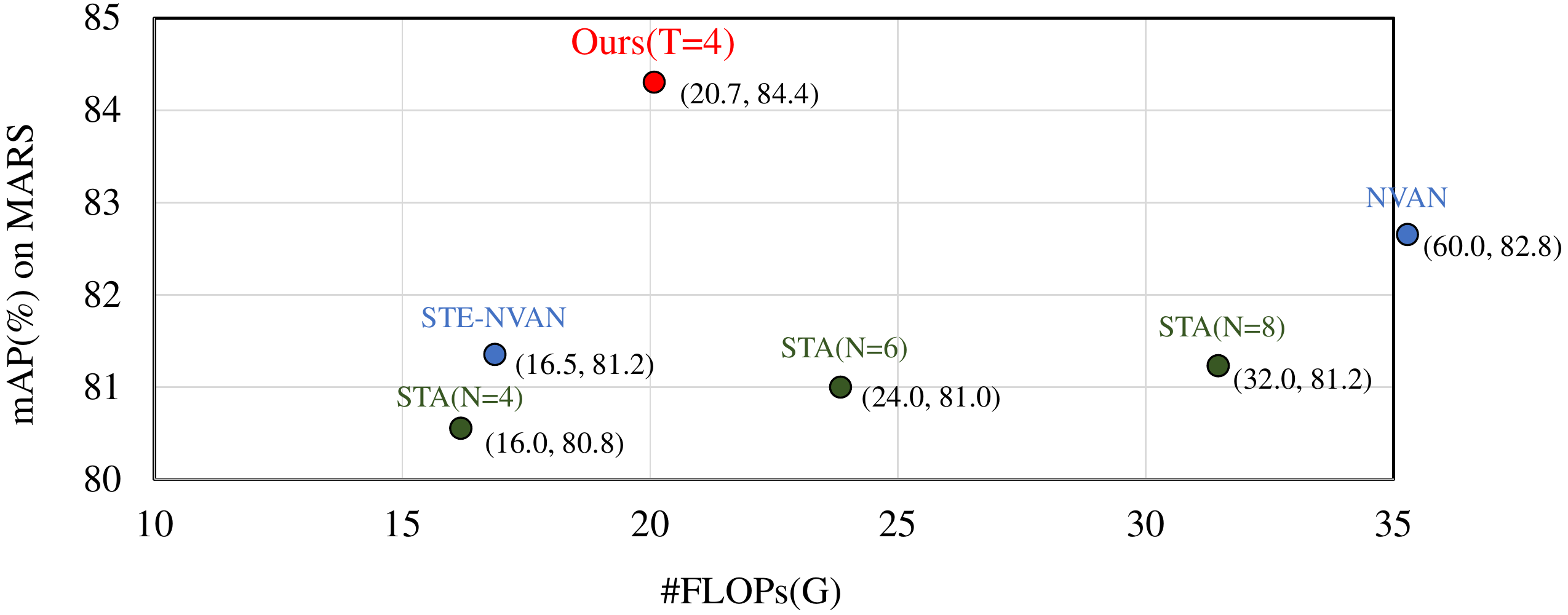} \\
	\caption{Computation-performance plot of our proposed multi-task model and other existing methods with attention mechanisms.}
	\label{fig:computing}
\end{figure*}
Besides the recognition performance, our model is extremely light-weighted. To take the computation complexity into consideration, we compare our method with existing methods that also use attention mechanisms on the performance-computation plot in Figure~\ref{fig:computing}. We visualize mAP on MARS dataset for the performance and FLOPs for computation counts. For STA~\cite{fu2019sta}, we follow the work in Liu et al.~\cite{LiuSpatially} which reports three variants of STA model with different numbers of sampled frames per sequence to better demonstrate their trade-off. The STE-NVAN and NVAN in Liu et al.~\cite{LiuSpatially} is also compared in this trade-off competition.

Since our model can achieve the best performance when the sampled frames number per sequence $T$ is set to 4, we directly use the best model to compare with other attentive models. As can be seen from Figure~\ref{fig:computing}, the results not only indicate the advantage of our model in both performance and computation counts, but also reveal the importance of using attribute in video-based Re-ID.

\subsection{Cross-dataset validation}
To evaluate the generalization performance of our multi-task model, we conduct the cross-dataset validation on the MARS (M) and DukeMTMC-VID (D) datasets. We train the baseline model and proposed method on the two datasets respectively, and test them on the other dataset. As shown in Table~\ref{table:cross}, the performance of two models decrease dramatically in cross-dataset test due to the domain difference. It is obvious that the performance decline of $M \to D$ test is less than the $D \to M$. A possible explanation is that, compared with DukeMTMC-VID, MARS contains more identities and track-lets. It is not surprising that larger training set can result in better models.

Although the proposed method doesn't show any significant advantages on domain adaptation in the cross-dataset validation, it still keeps absolute leading position in the Re-ID performance in every metric compared with baseline. It proves that the attribute embedding in the proposed method will not cause serious over-fitting.
\begin{table}[t]
	\caption{Cross-dataset validation (\%).}
	\centering
	\begin{tabular}{p{2cm}|c|c|c|c}
		\toprule
		\multirow{2}*{train $\to$ test } & \multicolumn{2}{c|}{Baseline.} & \multicolumn{2}{c}{proposed method}\\
		\cline{2-5}
		{}&mAP&R1&mAP&R1\\
		\cline{1-5}
		{$M \to M$}&79.4&84.9&\textbf{84.4}&\textbf{88.2} \\
		\cline{1-5}
		{$M \to D$}&47.3&49.1&\textbf{51.3}&\textbf{53.0} \\
		\cline{1-5}
		{$D \to D$}&91.6&92.2&\textbf{95.3}&\textbf{95.4} \\
		\cline{1-5}
		{$D \to M$}&24.0&40.3&\textbf{26.8}&\textbf{44.0} \\
		\bottomrule
	\end{tabular}
	\label{table:cross}
\end{table}

\section{Conclusion}
\label{sec:summary}

In this paper, we proposed a multi-task model which combines the Attribute-driven spatio-temporal Attention strategy and the Attribute-aware Identity-hard Triplet Loss to utilize the attribute information to help improve the video-based Re-ID performance. The Attribute-driven spatio-temporal Attention strategy could introduce the important spatio-temporal clues generated in the attribute recognition to encode the Re-ID features, while the Attribute-aware Identity-hard Triplet Loss could effectively reduce the Distance Variance among Different Positives. Both of these two strategies can highly improve the Re-ID performance, and the effectiveness of the proposed model is well demonstrated by experiments. 

Although effective, the proposed multi-task model does not make full use of the pose and motion attributes. They are simply combined with other ID-irrelevant attributes to form an overall attention. Since the track-lets obtained from two cameras are captured in a very short time, the walking direction, which appears as the pose in a camera, does have a strong relation. The movement and speed are also supposed to be consistent. What else, we found that some attributes like cloth color and gender contributes more than other attributes in the improvement of Re-ID performance in experiments, but addressing the influence of single attributes as well as their correlations with Re-ID is complicated, the page length of this paper does not allow a comprehensive study. So we planned to discuss these problems in future work.
\bibliographystyle{splncs04}
\bibliography{ref}
\end{document}